\documentclass[letterpaper, 10 pt, conference]{ieeeconf}  

\IEEEoverridecommandlockouts                              

\usepackage{multirow} 
\usepackage{graphicx}
\usepackage{epstopdf}
\usepackage{epsfig}
\usepackage{algorithm}
\usepackage{algorithmic}
\usepackage{amssymb}

\usepackage{amsmath}
\usepackage{amsthm}
\usepackage{fixmath}
\usepackage{caption}
\usepackage{hyperref}
\usepackage{subfigure}
\usepackage{cite}
\usepackage{tabularx}
\usepackage{afterpage}
\usepackage{booktabs}
\usepackage{array}
\usepackage{url}
\usepackage{soul}
\usepackage{xfrac}
\usepackage{float}
\usepackage{shuffle}
\usepackage{siunitx}
\usepackage[usenames,dvipsnames]{xcolor}
\usepackage{makecell}
\usepackage{arydshln}

\newcommand{\setcaptype}[1]{\def\@captype{#1}}

\title{\LARGE \bf
Planning for Dexterous Ungrasping: Secure Ungrasping through Dexterous Manipulation}

\author{Chung~Hee~Kim,
        Ka~Hei~Mak, 
        and~Jungwon~Seo 
\thanks{C. H. Kim, K. H. Mak, and J. Seo are with The Hong Kong University of Science and Technology (e-mail: chkimaa@connect.ust.hk; khmakac@connect.ust.hk; junseo@ust.hk). Corresponding author: J. Seo.}
}

\begin{document}

\maketitle
\thispagestyle{empty}
\pagestyle{empty}

\begin{abstract}
This paper presents a robotic manipulation technique for {\em dexterous ungrasping}. It refers to the capability of securely transferring a grasped object from the gripper to the robot's environment, i.e. the inverse of grasping or picking, through dexterous manipulation. The game of Go offers an example: consider how the player would typically place an initially pinch-grasped stone onto the board through the dexterous interaction between the fingers, the stone, and the board. Likewise, dexterous ungrasping addresses the necessity of changing the object's configuration relative to the gripper or the environment in order to securely keep hold of the object.
In particular, we present a planning framework for determining a feasible minimum-cost motion path that completes dexterous ungrasping. Digit asymmetry in a gripper, i.e. difference in digit lengths, is discovered as the key to feasible and secure ungrasping.
A set of experiments show the effectiveness of dexterous ungrasping in practical placement tasks.
\end{abstract}



\section{Introduction}
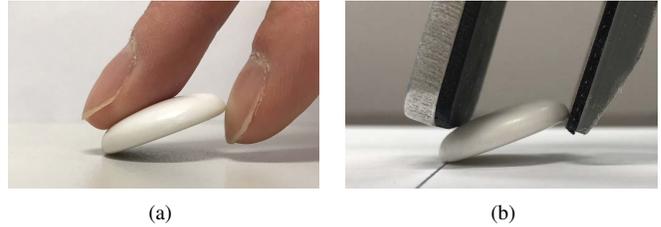
\begin{figure}[t]
\centering 
\includegraphics[width=\columnwidth]{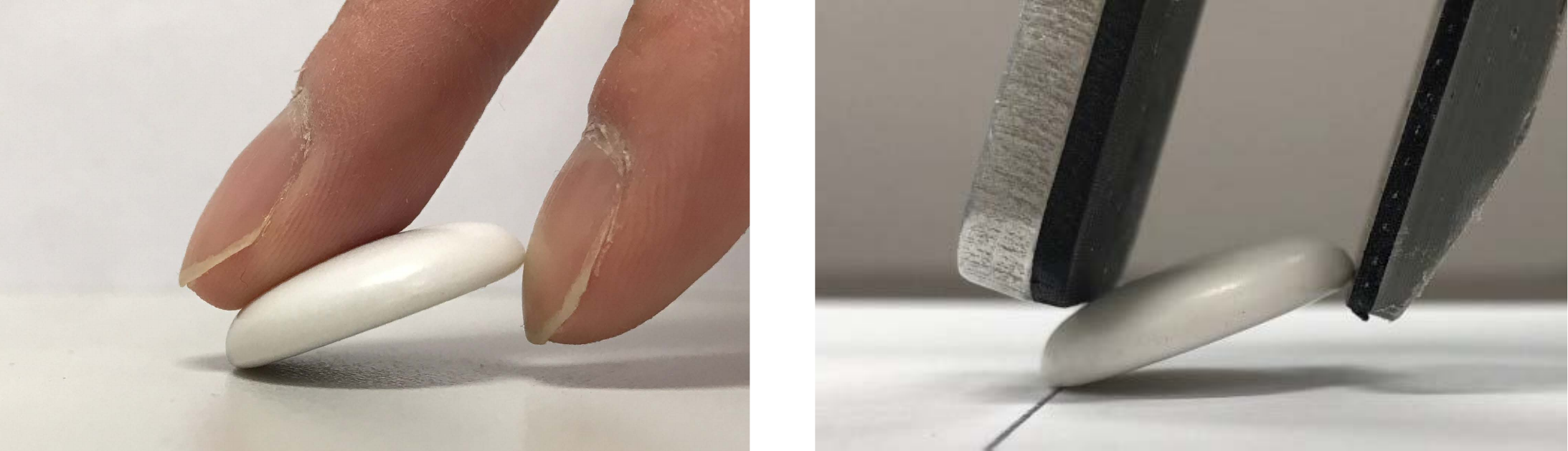}
\setlength{\unitlength}{1cm}
\begin{picture}(0,0)
\put(-2.45,0.0){\footnotesize (a)}
\put(2.1,0.0){\footnotesize (b)}
\end{picture}
\caption{
(a) Go stone placement by human. (b) How can a robot place a Go stone stably?}
\label{fig:go_stone_placement} 
\end{figure}

Consider the task of placing stones in the game of Go, as shown in Fig.~\ref{fig:go_stone_placement}. We humans seem to address this by exploiting in-hand dexterity, which we almost take for granted, to constantly reposition and reorient the object at hand while properly interacting with the environment. The current state of the art in robotic manipulation is, however, still some distance from practical proficiency in these capabilities that require dexterity; this may have necessitated the human operator that is actually able to place stones for AlphaGo in the matches against human champions. 

This work presents the technique of {\it dexterous ungrasping}, applicable to a range of object handling tasks such as Go stone placement. The term {\it un}grasping refers to the inverse of {\it grasping}; in other words, a robot ``passes'' an object to its environment throughout ungrasping. It is thus exemplified in placement, insertion, or assembly tasks. Grasping has been known as a challenging problem; likewise, ungrasping can be as difficult as grasping. Although there is a large body of work concerned with some particular ungrasping tasks (for example, peg-in-hole assembly), ungrasping through dexterous manipulation, the key novelty of the presented work, has received scant attention. Dexterity in ungrasping, which would not be necessary for simply dropping the object at gripper or for the traditional peg-in-hole assembly, is essential in case 
the object's pose needs to be carefully adjusted relative to the gripper and to an environment. This situation is particularly common in ungrasping slender or thin objects, which may begin with initial pinch grasping on the faces of the object and then involve a series of actions to relocate the finger from the face and to match the face with a target surface. An example can be seen in the way a Go player normally places a stone at a desired location through picking, in-hand manipulation, and interaction with the board using the index and the middle finger.

The key aspect of the presented work is planning for dexterous ungrasping. We provide a framework for planning complete, optimal, secure ungrasping motion. The planned path comprises motion primitives of contact interaction that can be realized with a simple two-digit, finger-thumb gripper model, one of the most common and versatile gripper architectures. The planner reveals that a structural asymmetry in the gripper, exemplified as the difference in digit lengths, is necessary to find a feasible ungrasping plan.
Our experiment results, featuring a range of scenarios including the task of Go stone placement, demonstrate that our approach leads to effective and secure dexterous ungrasping.

\section{Related Work}
\label{sec:related_work}

This study is closely related to robotic {\it dexterous manipulation}, also referred to as {\it in-hand manipulation}. It is generally defined as the capability of changing the configuration of a manipulated object relative to the gripper \cite{897777}. One classical approach to robotic dexterity is to devise ``dexterous hands'' with multi-jointed fingers and to manipulate the object at hand through finely controlled fingertip contacts. For example, the three-fingered Salisbury Hand \cite{mason1985robot} had nine degrees-of-freedom (DOF)---three DOF for each finger---which were shown to be the lower bound for performing dexterous manipulation with rigid, frictional, fixed (neither rolling nor sliding) contacts. In finger gaiting \cite{1087063, rus1999hand}, referring to changing grasps by repositioning fingers while the object is being manipulated by other fingers, it is often assumed that extra mobility is available, e.g. six DOF for each finger \cite{han1998dextrous}.

Alternatively, it is also possible to achieve advanced dexterity using grippers with relatively low-DOF. \cite{1087910} used a parallel-jaw gripper for regrasping, a sequential execution of grasping and ungrasping. A body of work \cite{trinkle1990planning, cole1992dynamic} shows that dexterous manipulation with simple grippers is facilitated by sliding contacts. Incorporating nonholonomic constraints by rolling contacts also renders simple grippers dexterous as studied in \cite{montana1988kinematics, 614264}. Grippers adopting active finger contact surfaces for exploiting rolling contacts have been presented in \cite{yuanicra2020}. Recent approaches to enhancing dexterity for simple grippers include taking advantage of external resources such as contact with environmental surfaces or dynamic motion \cite{6907062, 7913727} and incorporating mechanical compliance or underactuation \cite{deimel2016novel, doi:10.1177/0278364918802346}.

The manipulation for ungrasping in this work is directly applicable to part insertion/placement, which is important in many application domains. The classical peg-in-hole insertion, surveyed in \cite{mason2001mechanics}, in which the peg is push-inserted into the hole has received considerable attention from a range of aspects such as control strategies \cite{simunovic1975force} and mechanical design \cite{watson1978remote}. Our challenge tasks in this work are different from the traditional peg insertion in that dexterity is of critical importance. 
Robotic bin packing \cite{shome2019towards} is another relevant problem that addresses packing objects into a possibly confined space in an orderly manner. \cite{9384169} (and references therein) presents a learning-based approach to context-aware object placement.
Note, many practical solutions to insertion, placement, and packing usually feature custom-made automated systems, which are generally difficult to repurpose.

\section{Preliminaries}
\label{sec:ungrasping}

\begin{figure}[h]
\centering 
\includegraphics[width=\columnwidth]{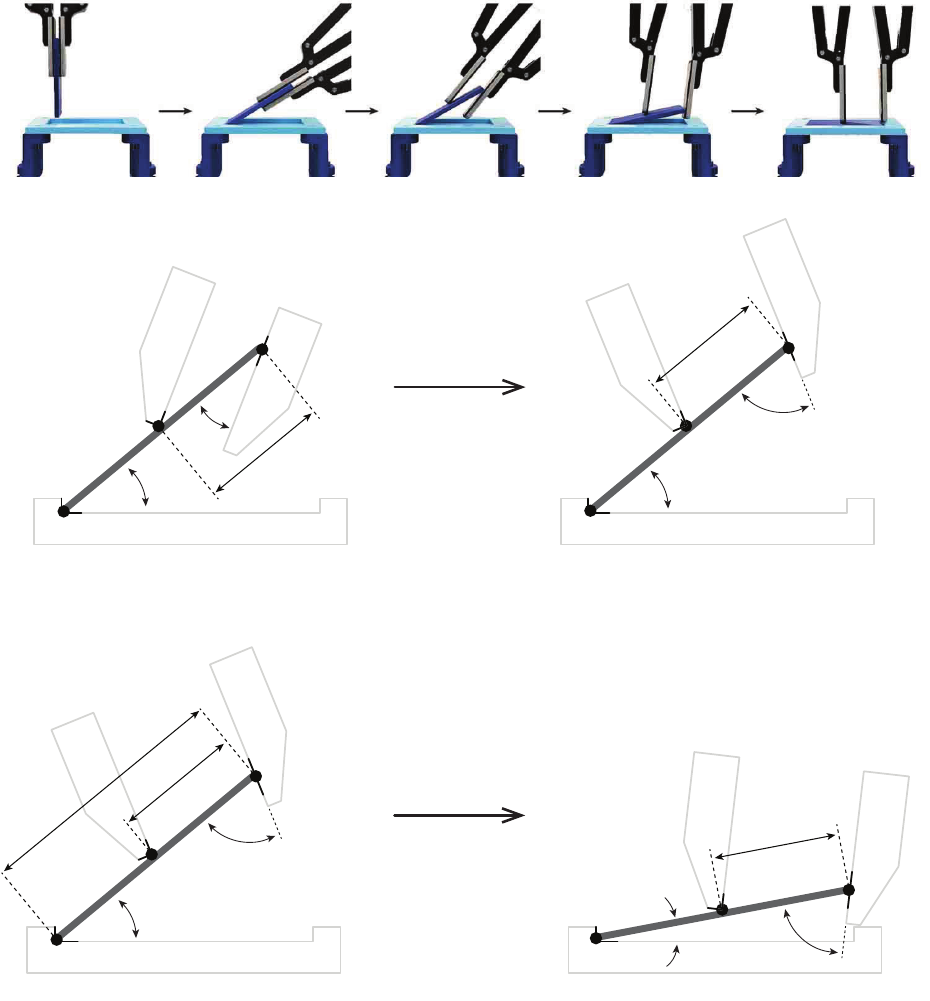}
\setlength{\unitlength}{1cm}
\begin{picture}(0,0)

\put(-0.3,7.7){\footnotesize (a)}
\put(-3.48,9.55){\scriptsize Parallel-finger gripper}
\put(-3.7,8.77){\scriptsize Object}
\put(-3.75,8.25){\scriptsize Hole}

\put(-0.3,4.1){\footnotesize (b)}
\put(-3.83,4.68){\footnotesize $G$}
\put(-2.95,5.2){\footnotesize $\theta$}
\put(-2.55,5.55){\footnotesize $\psi$}
\put(-1.9,5.3){\footnotesize $d_A$}
\put(-3.05,5.87){\footnotesize $A$}
\put(-3.1,7.0){\footnotesize Finger}
\put(-2.1,6.57){\footnotesize $B$}
\put(-1.65,6.45){\footnotesize Finger}
\put(-2.7,4.7){\footnotesize Hole}
\put(1.9,5.2){\footnotesize $\theta'$}
\put(2.7,5.6){\footnotesize $\psi'$}
\put(1.95,6.7){\footnotesize $d_A'$}

\put(1.1,4.68){\footnotesize $G$}
\put(2.05,5.47){\footnotesize $A$}

\put(3.07,6.4){\footnotesize $B$}

\put(-0.5,5.75){\footnotesize $\theta=\theta'$}
\put(-0.57,5.35){\footnotesize $\psi<\psi'$}
\put(-0.7,4.95){\footnotesize $d_A=d_A'$}

\put(-0.3,0.2){\footnotesize (c)}
\put(-3.9,.7){\footnotesize $G$}
\put(-3.,1.16){\footnotesize $\theta$}
\put(-2.25,1.65){\footnotesize $\psi$}
\put(-2.97,2.37){\footnotesize \rotatebox{42}{$d_A$}}
\put(-2.9,1.55){\footnotesize $A$}
\put(-1.92,2.5){\footnotesize $B$}
\put(-3.7,2.3){\footnotesize \rotatebox{42}{$\ell_\mathrm{obj}$}}
\put(-0.5,1.8){\footnotesize $\theta>\theta'$}
\put(-0.57,1.4){\footnotesize $\psi=\psi'$}
\put(-0.72,1.0){\footnotesize $d_A=d_A'$}

\put(1.7,1.4){\footnotesize $\theta'$}
\put(2.8,0.75){\footnotesize $\psi'$}
\put(2.7,2.){\footnotesize $d_A'$}
\put(1.1,.7){\footnotesize $G$}
\put(2.23,.96){\footnotesize $A$}
\put(3.6,1.4){\footnotesize $B$}

\end{picture}
\caption{(a) Shallow-depth insertion using a conventional parallel-finger gripper. (b) Releasing the linear object from the gripper: the contacts $G$ and $A$ remain fixed while $B$ is sliding and $\psi$ is increased. (c) Bringing the object to the hole: $G$, $A$, and $B$ remain fixed while $\theta$ is decreased.}
\label{fig:release_approach} 
\end{figure}
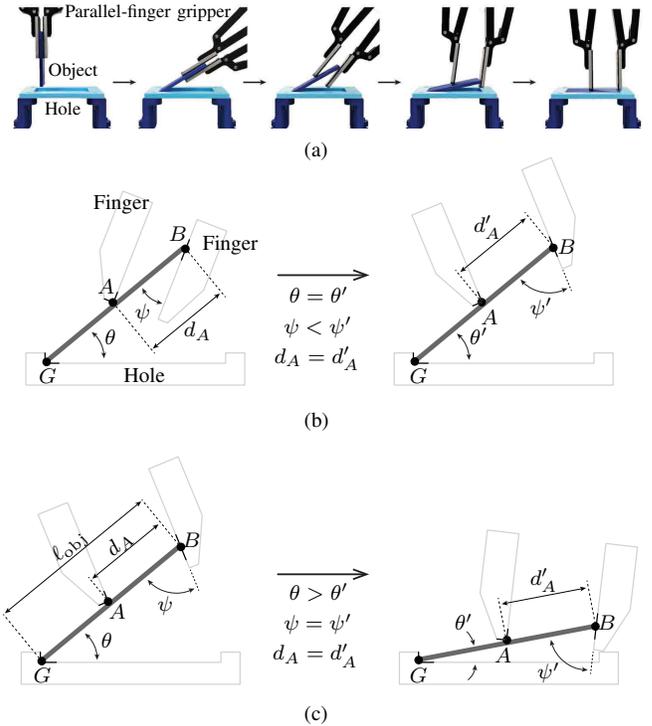

Our previous work on \textit{shallow-depth insertion} \cite{8598749}, depicted in Fig.~\ref{fig:release_approach}(a), verifies that it is possible to perform a certain type of dexterous ungrasping, i.e. dexterous insertion, securely without losing hold of the object. 
Here we review that work, but with a more detailed model of mechanics.

The key to successful shallow-depth insertion is the change of grasp realized by the motion primitive shown in Fig.~\ref{fig:release_approach}(b), modeling the real setting in Fig.~\ref{fig:release_approach}(a) on the plane normal to the hole's bottom face. Fig.~\ref{fig:release_approach}(b) thus features a grasp on a linear object composed of three point contacts at $G$, $A$, and $B$, formed by a flat-bottomed hole and a parallel-finger gripper with two fingers of the same length. This primitive is used for releasing the object from the gripper. At $B$, the finger slides on the object such that $B$ moves distally towards the fingertip. In the meantime, $G$ and $A$ are supposed to be fixed. It is kinematically possible to do so by coordinating the motions of the one-DOF gripper and a holonomic arm: the arm's mobility can be used for rolling (or fixing) the fingertip at $A$, counterclockwise as shown in Fig.~\ref{fig:release_approach}(b), and the gripper's mobility for providing enough space for the increasing distance between $A$ and the other finger's face.

\begin{figure}[h]
\centering 
\includegraphics[width=8cm]{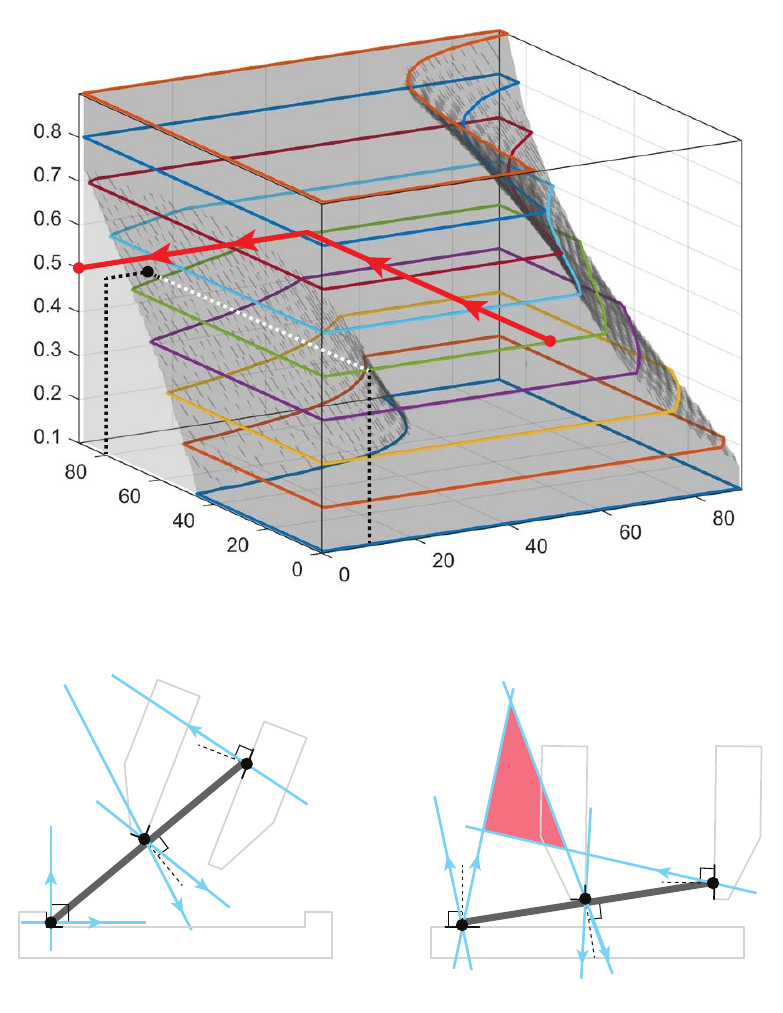}
\setlength{\unitlength}{1cm}
\begin{picture}(0,0)

\put(-4.4,3.9){\footnotesize (a)}
\put(-2.53,6.7){\footnotesize $\mathbf{q}_{\mathrm{init}}$}
\put(-7.5,7.75){\footnotesize \rotatebox{8}{$\mathbf{q}_{\mathrm{goal}}$}}
\put(-2.45,4.45){\footnotesize $\theta$ ($^\circ$)}
\put(-7.,4.6){\footnotesize $\psi$ ($^\circ$)}
\put(-8.26,7.3){\footnotesize $\delta_A$}
\put(-2.5,8.4){\footnotesize \rotatebox{8.}{$\mu_{A,B,G}=0.4$}}
\put(-6.9,7.25){\footnotesize (c)}

\put(-6.65,0.05){\footnotesize (b)}
\put(-8.0,1.6){\footnotesize $\mathbf{w}_{G1}$}
\put(-7.35,0.7){\footnotesize $\mathbf{w}_{G2}$}
\put(-7.0,1.88){\footnotesize $A$}
\put(-6.6,0.9){\footnotesize $\mathbf{w}_{A1}$}
\put(-6.2,1.1){\footnotesize $\mathbf{w}_{A2}$}
\put(-5.67,2.5){\footnotesize $B$}
\put(-6.75,3.05){\footnotesize $\mathbf{w}_{B}$}
\put(-7.83,0.6){\footnotesize $G$}

\put(-2.3,0.05){\footnotesize (c)}
\put(-3.6,0.6){\footnotesize $G$}
\put(-2.5,1.25){\footnotesize $A$}
\put(-0.92,1.33){\footnotesize $B$}
\put(-3.05,2.05){\footnotesize \boldmath$-$}

\put(-4.1,1.7){\footnotesize $\mathbf{w}_{G1}$}
\put(-3.45,1.65){\footnotesize $\mathbf{w}_{G2}$}
\put(-2.7,0.35){\footnotesize $\mathbf{w}_{A1}$}
\put(-2.1,0.35){\footnotesize $\mathbf{w}_{A2}$}
\put(-1.75,1.57){\footnotesize $\mathbf{w}_{B}$}

\end{picture}
\caption{(a) Complete insertion path (red arrow) contained in the set of force-closure grasps shaded gray (both light and dark), computed with the unit contact wrenches shown in (b), denoted $\mathbf{w}_{Xi}$ at the contact $X$. $\mu_X$ denotes the friction coefficient at $X$. The colored contours delineate the cross-sections of the set on the $\theta \psi$-planes. (c) Grasp with no force-closure, marked in the lightly shaded volume in (a), due to the flatness at $G$.}
\label{fig:pp_grasp_nopalm} 
\end{figure}
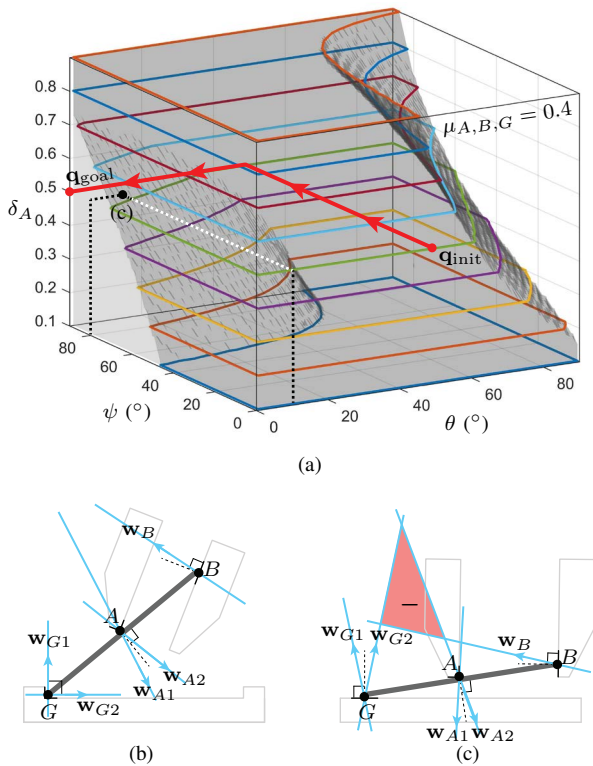

To complete the insertion, the motion primitive in Fig.~\ref{fig:release_approach}(b) is combined with another modeled in Fig.~\ref{fig:release_approach}(c), where the object is moved closer to the hole's bottom through a rotation about $G$, and all the contacts thus remain fixed. Fig.~\ref{fig:pp_grasp_nopalm}(a) shows an example motion path for complete shallow-depth insertion represented in the three-dimensional configuration space of the object-gripper-hole system, parametrized by $\theta$, $\psi$, and $d_A$ (refer to Fig.~\ref{fig:release_approach}). Note, the dimensionless form $\delta_A = \frac{d_A}{\ell_{\mathrm{obj}}}$ of $d_A$, where $\ell_{\mathrm{obj}}$ denotes the length of the linear object, was used in Fig.~\ref{fig:pp_grasp_nopalm}(a). At the terminal goal configuration of the path $\mathbf{q}_{\mathrm{goal}}$, $\theta$ is at its minimum, zero, such that the object's face is in contact with the hole's bottom face, and $\psi$ is at its maximum, $90^\circ$, such that the object is out of the workspace of the gripper, whose fingers are the same length. At the initial configuration $\mathbf{q}_{\mathrm{init}}$, the linear object is pinch-grasped by the fingers; thus, $\psi=0^\circ$. The motion primitive in Fig.~\ref{fig:release_approach}(b) (Fig.~\ref{fig:release_approach}(c)) reconfigures the system along the $\psi$-axis ($\theta$-axis), while keeping $\delta_A$ fixed. The entire motion from $\mathbf{q}_{\mathrm{init}}$ to $\mathbf{q}_{\mathrm{goal}}$ is therefore shown as a piecewise straight path on the $\theta\psi$-plane.

Throughout the entire insertion process, the object is kept in force-closure. In other words, the insertion motion path shown in Fig.~\ref{fig:pp_grasp_nopalm}(a) is contained in the volume representing the collection of force-closure grasps, shaded gray (regardless of color intensity for now), which is obtained in a numerical, sampling-based manner. For each sampled configuration $(\theta, \psi, \delta_A)$, a test for force-closure is formulated as a linear program verifying whether it is possible to span the entire wrench space \cite{Lynch:2017:MRM:3165183}. Fig.~\ref{fig:pp_grasp_nopalm}(b) shows the unit contact wrenches used for the test. These unit wrenches model the force interaction when the motion primitive in Fig.~\ref{fig:release_approach}(b) is executed. At $G$ and $A$, the contact wrenches are spanned by two unit wrenches, the normals of the hole corner edges at $G$ and the edges of the friction cone at $A$. The contact wrenches at $B$ are spanned solely by the edge of the friction cone that is consistent with the sliding of the object toward the fingertip. With these unit wrenches, the entire shaded volume represents the grasps in force-closure throughout the whole process, in a conservative manner because there will be an additional unit wrench at $B$, i.e. the other edge of the friction cone, when the other motion primitive shown in Fig.~\ref{fig:release_approach}(c) is executed. That the object is kept in force-closure enabled the successful insertion demonstrations in \cite{8598749}.

\section{Planning-through-Contact for Dexterous Ungrasping}
\label{sec:planning}

The viability of the special-case operation, shallow-depth insertion, motivates us to devise a generalized solution to dexterous ungrasping that can address a wider range of situations, e.g. placement onto a flat surface with no concavity, extending the task of insertion.
Critical issues expected in the generalization are as follows. First, consider a placement situation in Fig.~\ref{fig:pp_grasp_nopalm}(c) whereby the contact $G$ is on a flat surface, unlike Fig.~\ref{fig:pp_grasp_nopalm}(b) where $G$ is at a concave corner. The contact wrenches at $G$ are thus spanned by the friction cone edges. Here, force-closure is unattainable. The red-shaded region labeled with ``$-$'' represented according to the method of moment labeling \cite{mason2001mechanics} 
graphically indicates the inability to exhaust the entire wrench space.
This happens near the goal configuration, with relatively large $\psi$ and small $\theta$. The robot can thus lose the object near the goal (to be shown in our experiments). The collection of those configurations with no force-closure during the placement is represented as the volume shaded light gray in Fig.~\ref{fig:pp_grasp_nopalm}(a), around $\mathbf{q}_{\mathrm{goal}}$. Second, the two motion primitives used in the insertion (Fig.~\ref{fig:release_approach}(b-c)) are inadequate to fully navigate the configuration space, e.g. across different $\delta_A$ values in Fig.~\ref{fig:pp_grasp_nopalm}(a). This problem can complicate collision avoidance. This section presents a planning framework to address these concerns.

\begin{figure}[h!]
\centering 
\includegraphics[width=\columnwidth]{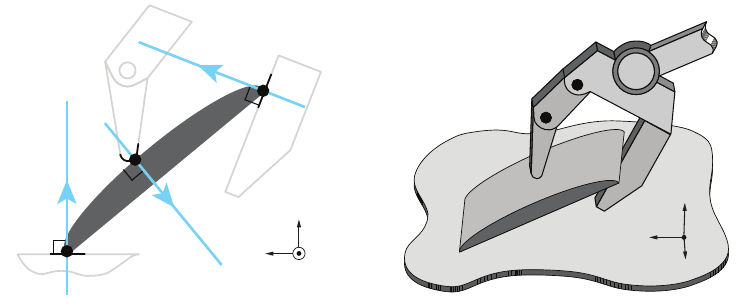}
\setlength{\unitlength}{1cm}
\begin{picture}(0,0)
\put(-2.25,0.2){\footnotesize (a)}
\put(-3.15,2.0){\footnotesize $A$}
\put(-1.2,2.87){\footnotesize $B$}
\put(-3.53,0.75){\footnotesize $G$}
\put(-1.05,2.25){\footnotesize Thumb}
\put(-3.,3.6){\footnotesize Finger}
\put(-1.45,0.97){\footnotesize $x$}
\put(-0.96,1.43){\footnotesize $z$}
\put(-0.96,0.78){\footnotesize $y$}

\put(2.0,0.2){\footnotesize (b)}
\put(1.0,1.4){\footnotesize \rotatebox{25}{Object}}
\put(1.2,3.0){\footnotesize Finger}
\put(3.2,1.9){\footnotesize Thumb}
\put(1.37,0.93){\footnotesize Environment}
\put(3.45,1.65){\footnotesize $z$}
\put(2.92,1.16){\footnotesize $x$}
\put(3.45,0.8){\footnotesize $y$}
\end{picture}
\caption{(a) Generalized model of planar ungrasping with a semi-elliptical object and a gripper whose fingertip (currently at $A$) can be placed freely relative to the thumb, as implied by the silhouette of a multi-jointed finger. The arrows are the contact normals. (b) 3D view of the ungrasping setting.}

\label{fig:local_geometry} 
\end{figure}
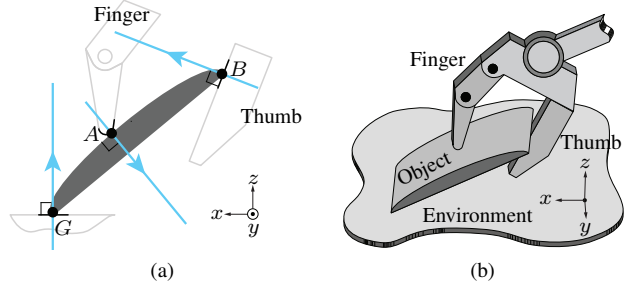

\subsection{Problem Formulation}
\label{subsec:problem_formulation}

We consider a generalized yet still planar ungrasping situation as shown in Fig.~\ref{fig:local_geometry}(a). Compared with the previous setting in Fig.~\ref{fig:release_approach}, it features a more sophisticated, finger-thumb gripper model in which the fingertip can be placed freely with respect to the thumb's face, a target surface without necessarily a concave corner, and an object with a convex contour such that the contact normals between the bodies are well-defined like Fig.~\ref{fig:local_geometry}(a). Similarly to the previous case, the gripper is assumed to be wielded by a three-DOF holonomic arm in the plane, and the contact interactions between the bodies in the object-gripper-environment system, all assumed to be rigid, are abstracted with the three unilateral point contacts at $G$ (between the object and the target surface), $A$ (between the fingertip and the object), and $B$ (between the object and the thumb's face).

Let $\mathcal{C}$ denote the configuration space of the object-gripper-environment system. The problem of the manipulation planning for secure ungrasping is then formalized as follows:
\begin{quote}
    {\em Find a feasible motion path in $\mathcal{C}$ along which the object initially pinched by the gripper is taken toward the goal configuration where it rests on a target surface, with no loss of grasp security all the way through.}
\end{quote}
\noindent
To keep hold of the object securely as required, $\mathcal{C}$ is limited to the configurations in which the contacts $G$, $A$, and $B$ are appropriately maintained. For example, at any time at least two contacts (or four unit contact wrenches) are necessary for force-closure in the plane \cite{Lynch:2017:MRM:3165183}. It is thus necessary to do planning-through-contact: how to arrange the contact interactions in the system, which will be embodied as dexterity, such that the object is safely ungrasped.

Each contact constraint at $G$, $A$, or $B$ can be regarded as a two-DOF joint in the sense that the contact can be either sliding or rolling. 
In case all the three contacts are active for example, the system is modeled as a planar linkage with three two-DOF joints, and a configuration of the system $\mathbf{q} \in \mathcal{C}$ is represented as a six-tuple. The use of the three parameters $\theta$, $\psi$, and $\delta_A$ in Fig.~\ref{fig:release_approach} is a specialization of this concept due to the kinematic constraints imposed by an ordinary parallel-finger gripper handling a linear object.

We make further remarks about the viability of our setting in Fig.~\ref{fig:local_geometry}(a). This planar setting abstracts a real-world ungrasping situation illustrated in Fig.~\ref{fig:local_geometry}(b) with a thin, flat object and a multi-jointed gripper, under the assumption that the object's motion out of the $xz$-plane can be effectively suppressed. To satisfy this, it is necessary to first align the object along an antipodal pair, for example, along a pair of parallel edges as shown in Fig.~\ref{fig:local_geometry}(b) with the rectangular object. This problem of regrasping/reorienting a pinch-grasped thin flat object has been a topic of interest in recent comprehensive studies \cite{7913727, doi:10.1177/0278364919880257}. These can be performed prior to our ungrasping operation for the initial alignment, and then the use of the planar setting is justified. 

\subsection{Planner Framework}
\label{subsec:planner_framework}
                                                                      
Algorithm 1 presents our solution to the planning-through-contact problem for dexterous ungrasping that is applicable to a high-dimensional space, facilitates constraint checking, and yields an optimal solution with respect to a criterion of choice. 
Overall, the algorithm is based on the RRT* algorithm \cite{doi:10.1177/0278364911406761}. The pseudocode thus features the key methods of RRT* for sampling the configuration space to build a space-filling tree $\mathcal{T}$ and for finding an optimal path from an initial configuration $\mathbf{q}_{\mathrm{init}}$ to a final $\mathbf{q}_{\mathrm{goal}}$. Specifically, we explore the search space through contact interactions, that is, sliding or rolling at the contacts $G$, $A$, and $B$, and adopt a test for grasp security when building the tree. As will be elaborated below, these make it possible for the RRT* variant to plan through contact for dexterous ungrasping. 

\begin{table}[h]
\centering
\small
\begin{tabularx}{\columnwidth}{l}
\toprule
\makecell[cl]{\textbf{Algorithm 1} Dexterous Ungrasping Planner} \\  
\midrule
\makecell[cl]{
\textbf{Input:} $\mathbf{q}_\mathrm{init}$, $\mathbf{q}_\mathrm{goal}$\\
\textbf{Output:} tree $\mathcal{T}$\\
{\footnotesize 1:} \hspace{1mm} $\mathcal{T}.\texttt{init}(\mathbf{q}_\mathrm{init})$ \\
{\footnotesize 2:} \hspace{1mm} \textbf{Repeat}\\
{\footnotesize 3:} \hspace{5mm} $\mathbf{q}_\mathrm{samp}\leftarrow \texttt{SampleDense}(\mathcal{C})$\\
{\footnotesize 4:} \hspace{5mm} $\mathbf{q}_\mathrm{nearest}\leftarrow \texttt{Nearest}(\mathcal{T}, \mathbf{q}_\mathrm{samp})$\\
{\footnotesize 5:} \hspace{5mm} $\mathbf{q}_\mathrm{new}\leftarrow \texttt{UngraspPrimitives}(\mathbf{q}_\mathrm{nearest}, \mathbf{q}_\mathrm{samp})$\\
{\footnotesize 6:} \hspace{5mm} \textbf{if} $\texttt{Free}(\mathbf{q}_\mathrm{nearest},\mathbf{q}_\mathrm{new})$ \textbf{then} \\
{\footnotesize 7:} \hspace{9mm} $\mathcal{T}.\texttt{InsertNode}(\mathbf{q}_\mathrm{new})$\\
{\footnotesize 8:} \hspace{9mm} $\mathcal{T}.\texttt{InsertEdge}(\mathbf{q}_\mathrm{nearest}, \mathbf{q}_\mathrm{new})$\\
{\footnotesize 9:} \hspace{9mm} $\mathcal{T}.\texttt{Rewire}$\\
{\footnotesize 10:} \hspace{0.1mm} \textbf{Until} $\mathcal{T}$ reaches $V(\mathbf{q}_\mathrm{goal})$, a neighborhood of $\mathbf{q}_\mathrm{goal}$\\
} \\ 
\bottomrule
\end{tabularx}
\end{table}

First, the search space is explored using the motion primitives that determine the standardized ways to traverse the space through contact interaction. Each motion primitive thus specifies the contact modes at $G$, $A$, and $B$. For example, the contact modes featured in the motion primitive in Fig.~\ref{fig:release_approach}(b)---rolling at $G$ and $A$, sliding at $B$---can be denoted by
\begin{equation}
    \label{eq:prim12}
    \mathsf{R}_G\mathsf{R}_A\mathsf{S}_B \quad \textrm{(Fig.~\ref{fig:release_approach}(b))}
\end{equation}
expressed as the sequence of the contact labels $\mathsf{R}_X$ (rolling at the contact $X$) or $\mathsf{S}_X$ (sliding at $X$). Likewise, the primitive in Fig.~\ref{fig:release_approach}(c) is denoted by
\begin{equation}
    \mathsf{R}_G\mathsf{R}_A\mathsf{R}_B \quad \textrm{(Fig.~\ref{fig:release_approach}(c))}
\end{equation}
From now on, motion primitives will be distinguished by the contact labels. In practice, a motion primitive also needs to specify the relationship between the configuration variables such that the search space can be traversed along a well-defined one-dimensional arc subject to the motion constraints imposed by the robot platform to use and the object's shape (example in Sec.~\ref{subsec:examples}).
The resulting path will then consist of the concatenation of the primitives. The method \texttt{UngraspPrimitives} in Algorithm 1 is the key to realize these ideas. It takes as input $\mathbf{q}_\mathrm{samp}$, a sampled configuration, and $\mathbf{q}_\mathrm{nearest}$, the node closest to $\mathbf{q}_\mathrm{samp}$ in the swath of $\mathcal{T}$. It then applies pre-defined motion primitives from $\mathbf{q}_\mathrm{nearest}$ to compute a set of locally reachable configurations. Among these, it returns $\mathbf{q}_\mathrm{new}$, the one closest to $\mathbf{q}_\mathrm{samp}$, to be potentially connected to $\mathcal{T}$ in the subsequent steps. The way we plan-through-contact with the motion primitives thus bears a similarity to the way differential constraints are typically addressed in sampling-based planning.

Second, we define the free configuration space $\mathcal{C}_{\mathrm{free}}$, i.e. the search space of the algorithm, as:
\begin{equation}
\mathcal{C}_{\mathrm{free}}=(\mathcal{C} \cap \mathcal{C}_{\mathrm{grasp}}) \setminus \mathcal{C}_{\mathrm{obs}}    
\end{equation}
where $\mathcal{C}_{\mathrm{obs}}$ is the obstacle space 
and $\mathcal{C}_{\mathrm{grasp}}$ is the collection of secure grasps. Therefore, not only kinematic collisions between the gripper and the environment but also grasp security is taken into consideration when an edge is added to $\mathcal{T}$. Specifically, in the \texttt{Free} method that runs a test to determine whether $\mathbf{q}_\mathrm{new}$ can be connected to $\mathcal{T}$, we characterize grasp security as force-closure. Given the motion primitive to reach $\mathbf{q}_\mathrm{new}$, force-closure is verified using a linear test transcribed with the contact modes determined by the primitive, as instanced in Sec.~\ref{sec:ungrasping}. When running Algorithm 1, the computational cost of the linear programming is thus incurred in addition to that of the baseline RRT*.

Examples of the cost function for the RRT*-based framework include
\begin{equation}
c(\mathbf{q})=c(p(\mathbf{q})) + c_\mathrm{add}(\mathbf{q},p(\mathbf{q}))
\end{equation}
where $c(\mathbf{q})$ denotes the scalar positive cost of a node $\mathbf{q}\in\mathcal{T}$ computed in an additive manner based on that of its parent $p(\mathbf{q})$.
We assign the added term $c_\mathrm{add}(\mathbf{q},p(\mathbf{q}))$ the number of contact mode switch-overs happening at $G$, $A$, and $B$ to reach $\mathbf{q}$ from $p(\mathbf{q})$, due to a possible change of motion primitive at $p(\mathbf{q})$. In other words, frequently changing motion primitives (i.e. contact modes) are penalized.
For example, if the motion primitive $\mathsf{R}_G\mathsf{R}_A\mathsf{S}_B$ is preceded by $\mathsf{R}_G\mathsf{R}_A\mathsf{R}_B$, $c_\mathrm{add}$ is assigned to be $1$ because there is one contact mode switch-over at $B$. 
The cost function is initially computed after $\mathbf{q}_\mathrm{new}$ is inserted into $\mathcal{T}$ (Line 7, Algorithm 1), and recalculated when $\mathcal{T}$ is rewired (Line 9). 

\subsection{Examples}
\label{subsec:examples}

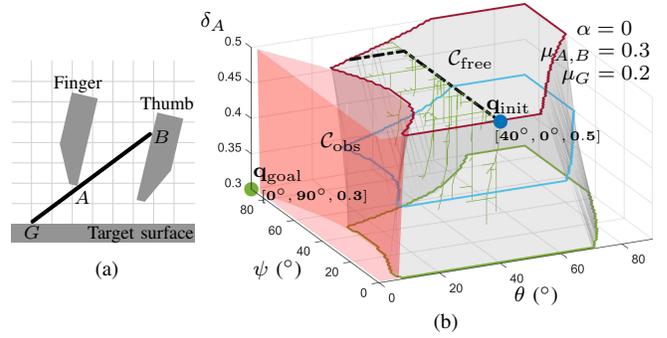
\begin{figure}[t]
\centering 
\includegraphics[width=\columnwidth]{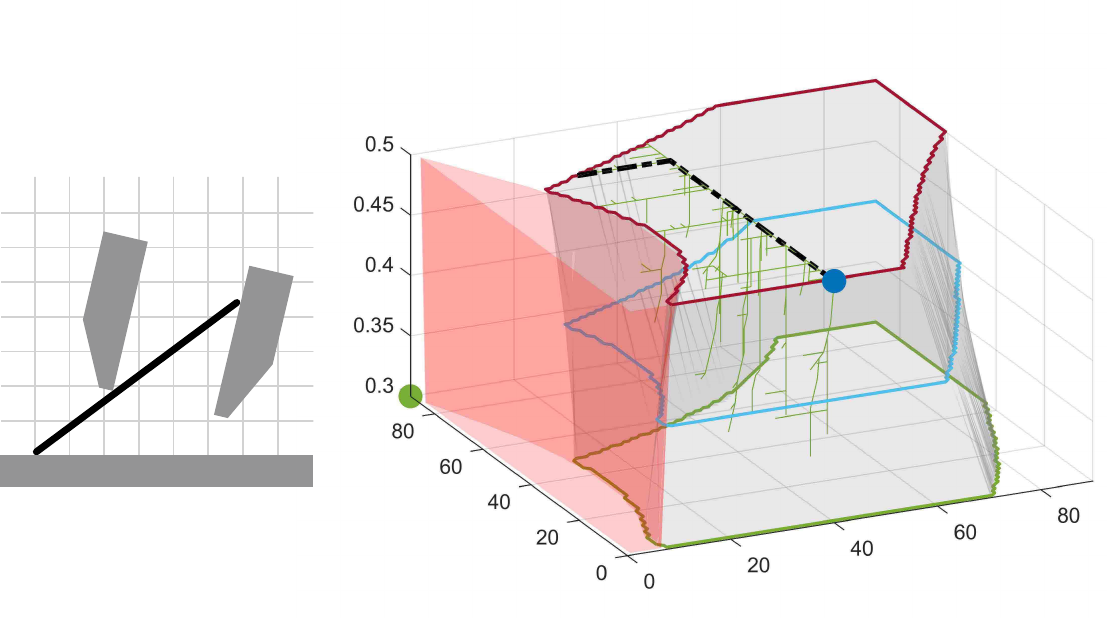}
\setlength{\unitlength}{1cm}
\begin{picture}(0,0)
\put(-3.2,1.0){\footnotesize (a)}
\put(-4.15,1.48){\scriptsize $G$}
\put(-3.48,1.95){\scriptsize $A$}
\put(-2.43,2.75){\scriptsize $B$}
\put(-3.3,1.48){\scriptsize Target surface}
\put(-3.75,3.48){\scriptsize Finger}
\put(-2.6,3.2){\scriptsize Thumb}

\put(1.3,0.3){\footnotesize (b)}
\put(2.37,0.65){\footnotesize $\theta$ ($^\circ$)}
\put(-1.1,1.0){\footnotesize $\psi$ ($^\circ$)}
\put(-1.8,4.3){\footnotesize $\delta_A$}
\put(2.0,3.2){\footnotesize $\mathbf{q}_\mathrm{init}$}
\put(-1.1,2.3){\footnotesize $\mathbf{q}_\mathrm{goal}$}
\put(1.5,3.8){\footnotesize $\mathcal{C}_\mathrm{free}$}
\put(-0.2,2.7){\footnotesize $\mathcal{C}_\mathrm{obs}$}
\put(3.2,4.20){\footnotesize $\alpha=0$} 
\put(2.72,3.90){\footnotesize $\mu_{A,B}=0.3$}
\put(3.0,3.60){\footnotesize $\mu_G=0.2$}
\put(2.1,2.8){\tiny {$\mathbf{[40^\circ, 0^\circ, 0.5]}$}}
\put(-1.0,2.0){\tiny $\mathbf{[0^\circ, 90^\circ, 0.3]}$}

\end{picture}
\caption{(a) Planning scene for the placement of a linear object with a parallel-finger gripper. The collision hulls of the digits and the target surface are as depicted. (b) Failed to connect $\mathbf{q}_\mathrm{init}$ and $\mathbf{q}_\mathrm{goal}$. $\mathcal{C}_\mathrm{free}$ ($\mathcal{C}_\mathrm{obs}$) is the region shaded gray (red).}
\label{fig:failure_example} 
\end{figure}

We now present numerical examples of planned dexterous ungrasping generated by our software\footnote{\url{https://github.com/HKUST-RML/dexterous\_ungrasping\_planner}} implementing Algorithm 1. Our planning scenes feature a two-fingered gripper interacting with a flat-bottomed object and other environmental surfaces (see Fig.~\ref{fig:failure_example}(a) for example). The gripper is arranged to be a one-DOF parallel-finger gripper. Its two flat-faced fingers are controlled to move in the direction normal to the face. 
This arrangement is not only practically important but also conducive to visualizing the planning results, again using the three parameters $\theta$, $\psi$, and $\delta_A$.
Note, all the three contacts $G$, $A$, an $B$ are kept active all the way. At the initial configuration $\mathbf{q}_{\mathrm{init}}$ where the object is in a stable pinch grasp, 
$\psi$ is zero. At the goal $\mathbf{q}_{\mathrm{goal}}$,
$\theta$ is zero (object lying on the target surface) and $B$ is at the tip of the thumb (object almost released from the gripper).

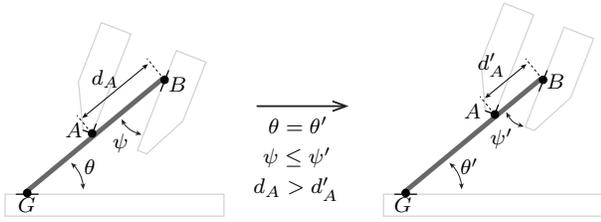
\begin{figure}[t]
\centering 
\includegraphics[width=\columnwidth]{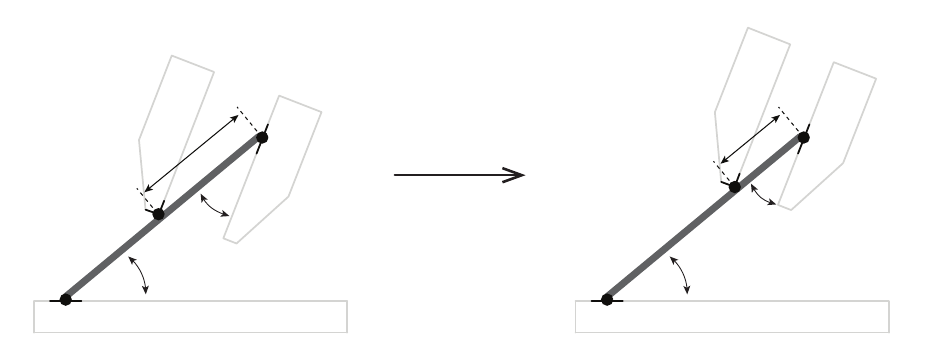}
\setlength{\unitlength}{1cm}
\begin{picture}(0,0)
\put(-3.83,0.6){\footnotesize $G$}
\put(-2.95,1.1){\footnotesize $\theta$}
\put(-2.57,1.45){\footnotesize $\psi$}
\put(-2.85,2.3){\footnotesize $d_A$}
\put(-3.2,1.6){\footnotesize $A$}
\put(-1.83,2.2){\footnotesize $B$}
\put(-0.5,1.65){\footnotesize $\theta=\theta'$}
\put(-0.57,1.25){\footnotesize $\psi\leq\psi'$}
\put(-0.7,0.85){\footnotesize $d_A>d_A'$}
\put(2.05,1.1){\footnotesize $\theta'$}
\put(2.45,1.5){\footnotesize $\psi'$}
\put(2.28,2.5){\footnotesize $d_A'$}
\put(1.17,0.6){\footnotesize $G$}
\put(2.1,1.85){\footnotesize $A$}
\put(3.2,2.25){\footnotesize $B$}
\end{picture}
\caption{Motion primitive to decrease $d_A$. Both $A$ and $B$ slide while $G$ remains fixed. 
}
\label{fig:gen_release} 
\end{figure}

The set of motion primitives used in the examples include the two, $\mathsf{R}_G\mathsf{R}_A\mathsf{S}_B$ and $\mathsf{R}_G\mathsf{R}_A\mathsf{R}_B$, in Fig.~\ref{fig:release_approach}(b-c). In case of object shape as a line segment, the former one $\mathsf{R}_G\mathsf{R}_A\mathsf{S}_B$ is implemented using the parallel-finger gripper by imposing the following motion constraints
\begin{equation}
    \label{eq:gripper}
    \tau_{\rm gripper} = \dot{(d_A \sin \psi)}=d_A \dot{\psi} \cos \psi
\end{equation}
where $\tau_{\rm gripper}$ denotes the commanded opening speed of the gripper in terms of the gripper's orientation $\psi$ relative to the object. At the same time, the arm is controlled to increase $\psi$. Given a different object shape, the control law can be adapted accordingly. In the latter, $\mathsf{R}_G\mathsf{R}_A\mathsf{R}_B$, only the gripper's orientation $\theta$ relative to the surface is changed by the arm. In addition, the motion primitive illustrated in Fig.~\ref{fig:gen_release} with contact labels
\begin{equation}
    \mathsf{R}_G\mathsf{S}_A\mathsf{S}_B \quad \textrm{(Fig.~\ref{fig:gen_release})}
\end{equation}
is applied to changing $\delta_A$ by sliding $A$ and $B$ at the same time. This is also executed by the gripper control law Eq.~\ref{eq:gripper} while the arm is controlled to move $A$, in case of a linear object. The three primitives enable the planner to fully explore the three-dimensional configuration space $(\theta,\psi,\delta_A)$.

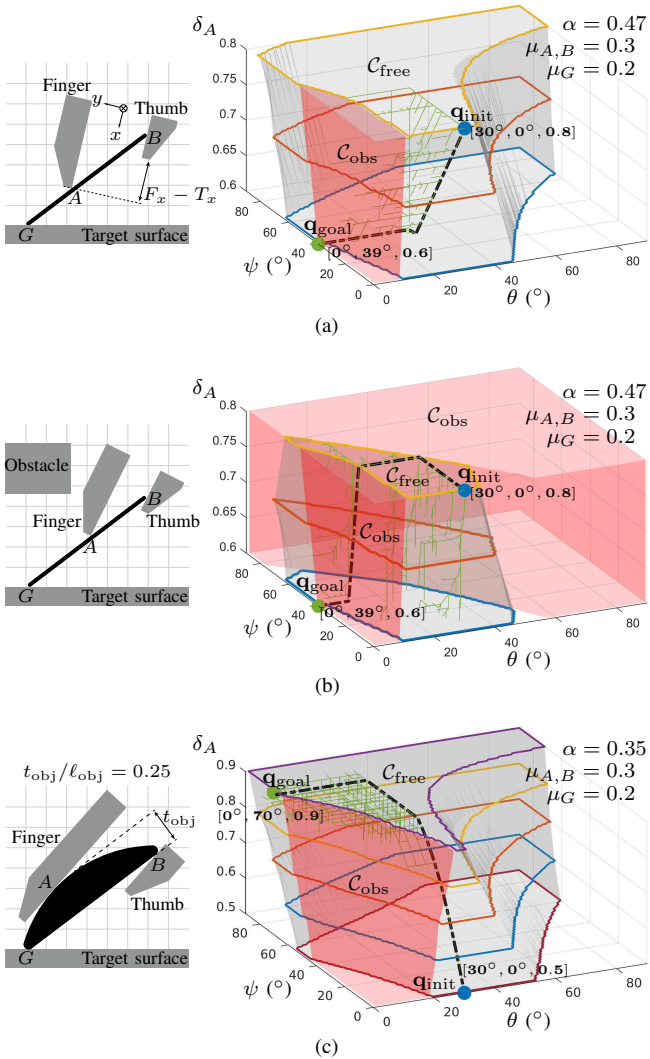
\begin{figure}[h]
\centering 
\includegraphics[width=\columnwidth]{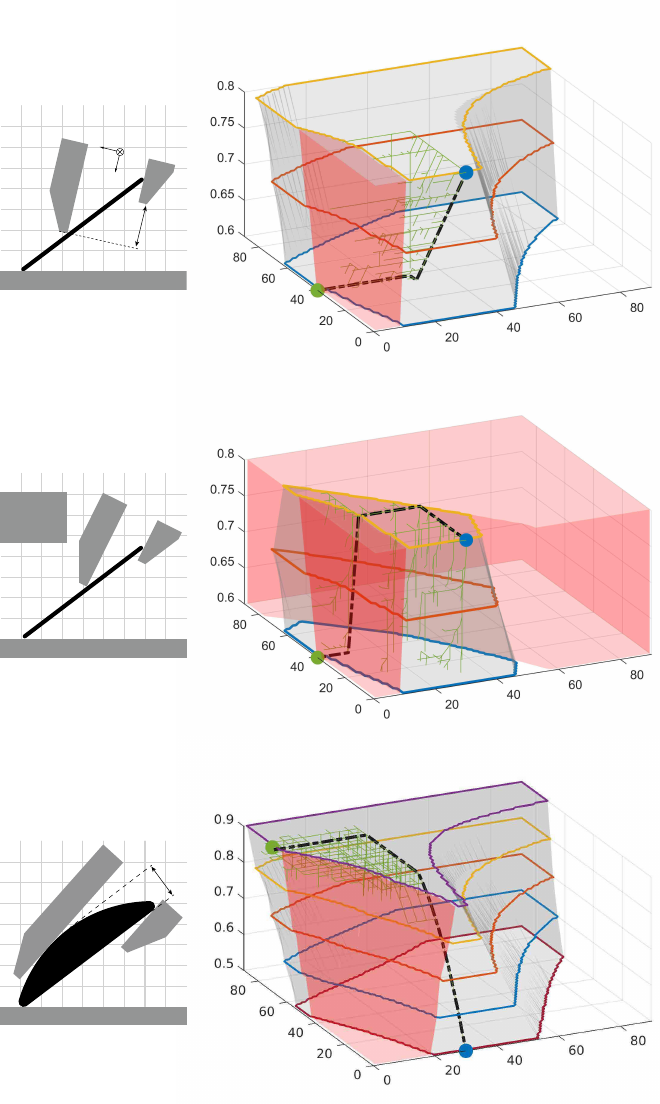}
\setlength{\unitlength}{1cm}
\begin{picture}(0,0)
\put(-0.2,9.9){\footnotesize (a)}
\put(-2.49,11.65){\scriptsize $F_x - T_x$}
\put(-4.15,11.1){\scriptsize $G$}
\put(-3.5,11.6){\scriptsize $A$}
\put(-2.45,12.4){\scriptsize $B$}
\put(-3.3,11.1){\scriptsize Target surface}
\put(-3.82,13.1){\scriptsize Finger}
\put(-2.6,12.8){\scriptsize Thumb}
\put(-2.93,12.45){\scriptsize $x$}
\put(-3.17,12.94){\scriptsize $y$}
\put(2.35,10.27){\footnotesize $\theta$ ($^\circ$)}
\put(-1.15,10.7){\footnotesize $\psi$ ($^\circ$)}
\put(-1.83,13.9){\footnotesize $\delta_A$}
\put(1.6,12.75){\footnotesize $\mathbf{q}_\mathrm{init}$}
\put(-0.35,11.25){\footnotesize $\mathbf{q}_\mathrm{goal}$}
\put(0.5,13.35){\footnotesize $\mathcal{C}_\mathrm{free}$}
\put(0.1,12.2){\footnotesize $\mathcal{C}_\mathrm{obs}$}
\put(1.85,12.5){\tiny {$\mathbf{[30^\circ, 0^\circ, 0.8]}$}}
\put(-0.05,10.9){\tiny $\mathbf{[0^\circ, 39^\circ, 0.6]}$}
\put(3.09,13.92){\footnotesize $\alpha=0.47$}
\put(2.6,13.62){\footnotesize $\mu_{A,B}=0.3$}
\put(2.88,13.32){\footnotesize $\mu_G=0.2$}

\put(-0.2,5.1){\footnotesize (b)}
\put(-4.15,6.3){\scriptsize $G$}
\put(-3.3,6.95){\scriptsize $A$}
\put(-2.45,7.6){\scriptsize $B$}
\put(-3.3,6.3){\scriptsize Target surface}
\put(-3.95,7.3){\scriptsize Finger}
\put(-2.44,7.3){\scriptsize Thumb}
\put(-4.33,8.05){\scriptsize Obstacle}
\put(2.35,5.45){\footnotesize $\theta$ ($^\circ$)}
\put(-1.15,5.88){\footnotesize $\psi$ ($^\circ$)}
\put(-1.83,9.07){\footnotesize $\delta_A$}
\put(1.7,7.95){\footnotesize $\mathbf{q}_\mathrm{init}$}
\put(-0.45,6.5){\footnotesize $\mathbf{q}_\mathrm{goal}$}
\put(0.4,7.15){\footnotesize $\mathcal{C}_\mathrm{obs}$}
\put(0.75,7.9){\footnotesize $\mathcal{C}_\mathrm{free}$}
\put(1.3,8.7){\footnotesize $\mathcal{C}_\mathrm{obs}$}
\put(1.85,7.7){\tiny $\mathbf{[30^\circ, 0^\circ, 0.8]}$}
\put(-0.15,6.1){\tiny $\mathbf{[0^\circ, 39^\circ, 0.6]}$}
\put(3.09,9.02){\footnotesize $\alpha=0.47$}
\put(2.6,8.72){\footnotesize $\mu_{A,B}=0.3$}
\put(2.88,8.42){\footnotesize $\mu_G=0.2$}

\put(-0.2,0.3){\footnotesize (c)}
\put(-4.15,1.49){\scriptsize $G$}
\put(-3.9,2.5){\scriptsize $A$}
\put(-2.4,2.73){\scriptsize $B$}
\put(-2.23,3.38){\scriptsize $t_\mathrm{obj}$}
\put(-4.08,3.97){\scriptsize $t_\mathrm{obj}/\ell_\mathrm{obj}=0.25$}
\put(-3.3,1.49){\scriptsize Target surface}
\put(-4.23,3.1){\scriptsize Finger}
\put(-2.65,2.2){\scriptsize Thumb}

\put(2.35,0.65){\footnotesize $\theta$ ($^\circ$)}
\put(-1.15,1.07){\footnotesize $\psi$ ($^\circ$)}
\put(-1.83,4.34){\footnotesize $\delta_A$}
\put(1.1,1.2){\footnotesize $\mathbf{q}_\mathrm{init}$}
\put(-0.9,3.9){\footnotesize $\mathbf{q}_\mathrm{goal}$}
\put(0.7,3.95){\footnotesize $\mathcal{C}_\mathrm{free}$}
\put(0.25,2.47){\footnotesize $\mathcal{C}_\mathrm{obs}$}
\put(3.09,4.27){\footnotesize $\alpha=0.35$} 
\put(2.6,3.97){\footnotesize $\mu_{A,B}=0.3$}
\put(2.88,3.67){\footnotesize $\mu_G=0.2$}
\put(1.75,1.35){\tiny $\mathbf{[30^\circ, 0^\circ, 0.5]}$}
\put(-1.5,3.4){\tiny $\mathbf{[0^\circ, 70^\circ, 0.9]}$}

\end{picture}
\caption{Successful planning for dexterous ungrasping (a) with digit asymmetry, (b) around another obstacle, and (c) with a semi-elliptical object. Each example depicts the planning scene on the left with the collision hulls of the bodies that can interact. The planned paths connecting $\mathbf{q}_\mathrm{init}$ and $\mathbf{q}_\mathrm{goal}$ are shown in the $\theta\psi\delta_A$-space. 
$\mathcal{C}_\mathrm{free}$ ($\mathcal{C}_\mathrm{obs}$) is the region shaded gray (red). The green branches represent the search tree $\mathcal{T}$, constructed through 1,000 iterations (this took 36-38 seconds with our software written in \textsc{Matlab}).}
\label{fig:successful_examples} 
\end{figure}

\subsubsection{Failure with an ordinary parallel-finger gripper}

First, we ran the planner with an ordinary parallel-finger gripper whose fingers are the same length (Fig.~\ref{fig:failure_example}(a)). The planner was unable to find a feasible ungrasping path for the linear object because $\mathbf{q}_{\mathrm{goal}}$ lies outside $\mathcal{C}_{\mathrm{free}}$ even when all the contacts $G$, $A$, and $B$ are fixed, as plotted in Fig.~\ref{fig:failure_example}(b). In fact, all the other potential goal configurations at $\psi=90^\circ$ and $\theta=0^\circ$ lie outside $\mathcal{C}_{\mathrm{free}}$, as aforementioned (recall Fig.~\ref{fig:pp_grasp_nopalm}(c)).

\subsubsection{Successful planning with digit asymmetry}

Next, we seek to exploit the increased freedom of the gripper model in Fig.~\ref{fig:local_geometry} by considering an asymmetric parallel-finger gripper. See the planning scene in Fig.~\ref{fig:successful_examples}(a) again with a linear object. The two flat-faced fingers are now different lengths (i.e. asymmetric) although these still open/close in the direction normal to the face. We call the longer one a finger and the shorter one a thumb. The asymmetry is quantified using the following dimensionless parameter $\alpha$:
\begin{equation}
    \alpha=\frac{F_x - T_x}{\ell_{\mathrm{obj}}}
\end{equation}
where $F_x$ ($T_x$) denotes the position of the finger's tip (thumb's tip) along the $x$-direction parallel to its contact surface (Fig.~\ref{fig:successful_examples}(a)). $\alpha$ is always nonnegative since $F_x \geq T_x$, i.e. the finger is no shorter than the thumb.
The value of asymmetry $\alpha$ dictates the value of $\psi$ at $\mathbf{q}_{\mathrm{goal}}$ as follows: 
\begin{equation}
    \label{eqn:psi_goal}
    \psi_{\mathrm{goal}} = \arccos \left( \frac{\alpha}{\delta_{A}} \right)
\end{equation}
$\alpha$ and $\delta_A$ are also evaluated at $\mathbf{q}_{\mathrm{goal}}$, where $B$ is at the thumb's tip. For an ordinary parallel-finger gripper with same-length fingers, $\alpha$ is kept zero and thus $\psi_\mathrm{goal}=90^\circ$. For a gripper with positive $\alpha$, $\psi_\mathrm{goal}$ becomes less than $90^\circ$. This enables the planner to find a complete solution. Given $\alpha=0.47$, a path to $\mathbf{q}_{\mathrm{goal}}$ where $\psi_\mathrm{goal}=39^\circ$ and $\delta_A=0.6$ was found in $\mathcal{C}_\mathrm{free}$ as shown in Fig.~\ref{fig:successful_examples}(a). The path, depicted as the black dashed line in the figure, represents a sequence of three motion primitives, each of which is shown as a line segment of fixed orientation: $\mathsf{R}_G\mathsf{S}_A\mathsf{S}_B$ at the beginning, then $\mathsf{R}_G\mathsf{R}_A\mathsf{S}_B$ briefly, and finally $\mathsf{R}_G\mathsf{R}_A\mathsf{R}_B$. 

\subsubsection{Successful planning around another obstacle}
In Fig.~\ref{fig:successful_examples}(b), a fixed obstacle was added to the planning scene, which resulted in a greatly enlarged $\mathcal{C}_\mathrm{obs}$. With the same asymmetric gripper setting as the previous example, our planner found a solution, a sequence of four motion primitives.

\subsubsection{Successful planning with semi-elliptical object}
Fig.~\ref{fig:successful_examples}(c) features a planning scene with a nonlinear, semi-elliptical object. The path was planned with two primitives, $\mathsf{R}_G\mathsf{R}_A\mathsf{S}_B$ and $\mathsf{R}_G\mathsf{R}_A\mathsf{R}_B$. It was necessary to adapt the $\mathsf{R}_G\mathsf{R}_A\mathsf{S}_B$ primitive to two distinct cases. First, when $A$ is formed in the interior of the finger's face, the finger rolls on the curved object. Second, when $A$ is at the sharp fingertip, the finger rotates about it and the location of $A$ is fixed. The planned path begins with $\mathsf{R}_G\mathsf{R}_A\mathsf{S}_B$ with $A$ on the finger's face, switches to another $\mathsf{R}_G\mathsf{R}_A\mathsf{S}_B$ with $A$ at the fingertip, and terminates with $\mathsf{R}_G\mathsf{R}_A\mathsf{R}_B$.

\textbf{\textit{Remark}}: These examples indicate the advantages offered by the introduction of digit asymmetry, i.e. the combination of a longer finger and a shorter thumb, enabled by the generalized gripper model (Fig.~\ref{fig:local_geometry}). First, it becomes possible to maintain force-closure all the way through. This is because digit asymmetry provides a greater degree of flexibility in the way ungrasping can be terminated: $\psi_{\mathrm{goal}}$ can be set less than $90^\circ$ as discussed. Second, decreased $\psi_{\mathrm{goal}}$ will reduce the duration of ungrasping. Third, the component of $\mathcal{C}_{\mathrm{obs}}$ representing the collision between the thumb and the target surface shrinks with increased $\alpha$ because of the relatively shortened thumb (compare Fig.~\ref{fig:failure_example} and Fig.~\ref{fig:successful_examples}(a)). Practically, these benefits can easily be reaped by using the asymmetric parallel-finger gripper featured in the examples.

\section{Experiments}
\label{sec:experiments}

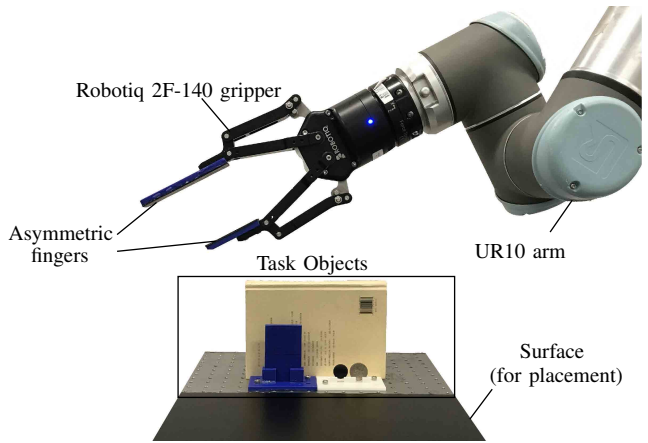
\begin{figure}[t]
\centering 
\includegraphics[width=\columnwidth]{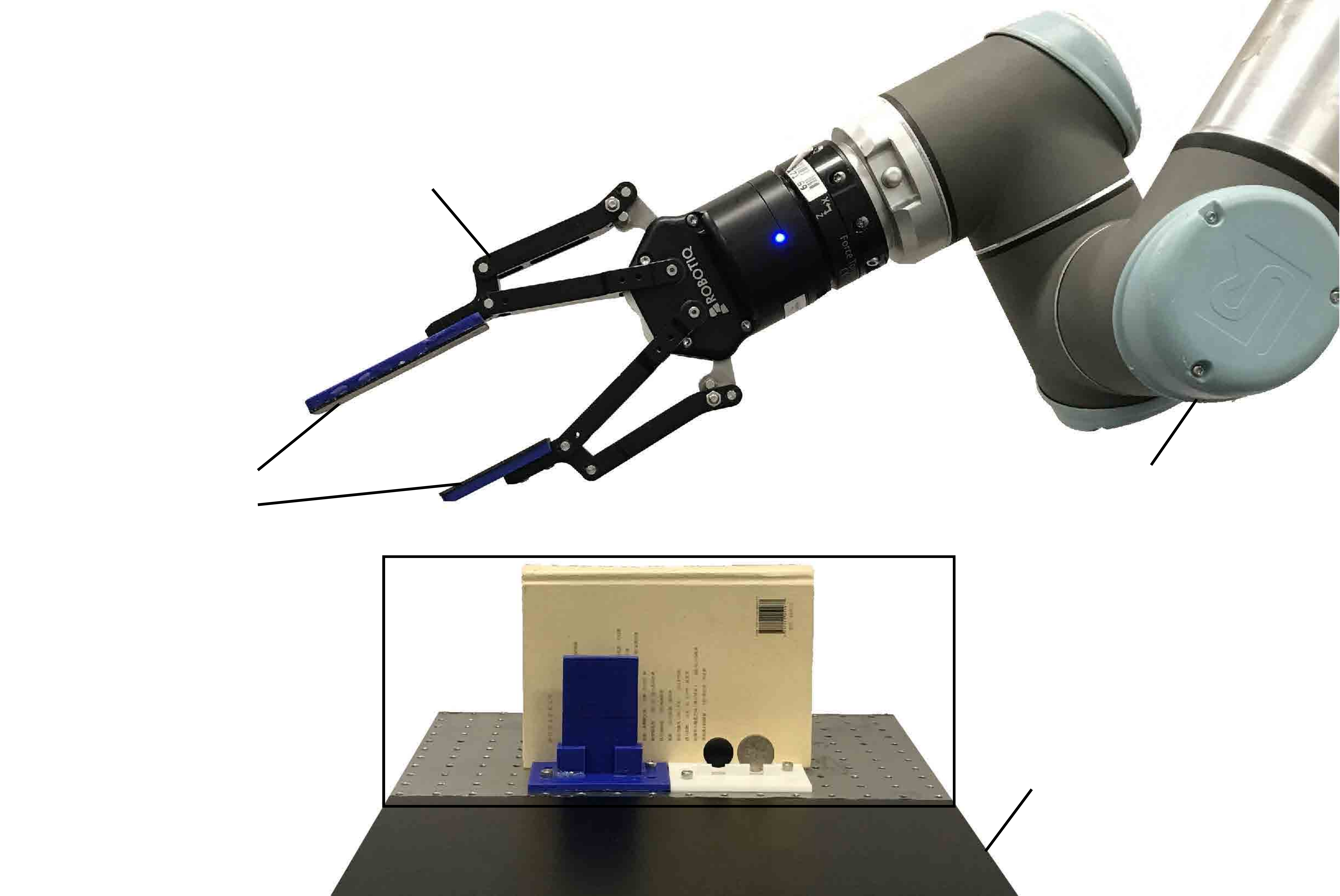}
\setlength{\unitlength}{1cm}
\begin{picture}(0,0)
\put(2.1,2.85){\footnotesize UR10 arm}
\put(-3.2,5.0){\footnotesize Robotiq 2F-140 gripper}
\put(2.7,1.53){\footnotesize Surface}
\put(2.3,1.2){\footnotesize (for placement)}
\put(-0.8,2.7){\footnotesize Task Objects}
\put(-4.1,3.1){\footnotesize Asymmetric}
\put(-3.8,2.8){\footnotesize fingers}
\end{picture}
\caption{Our experiment setting featuring a gripper with asymmetric fingers and a UR10 arm. Also shown are test objects: a 3D-printed plate, a hardcover book, a Go stone, and a coin.}
\label{fig:exp_setting} 
\end{figure}

This section presents a set of experiments in dexterous ungrasping. Specifically, we examined the performance of precision placement: an object is to be placed at a specific location on a flat tabletop while being held securely all the way through. This is more challenging than insertion tasks (recall Fig.~\ref{fig:pp_grasp_nopalm}(b-c)). The objects we tested include a coin, a hardcover book, a Go stone, and a 3D-printed rectangular plate. Fig.~\ref{fig:exp_setting} shows our setting with a Robotiq 2F-140 gripper carried by a UR10 arm. The gripper is equipped with customized 3D-printed fingers that vary in lengths to realize necessary digit asymmetry with different $\alpha$ values.

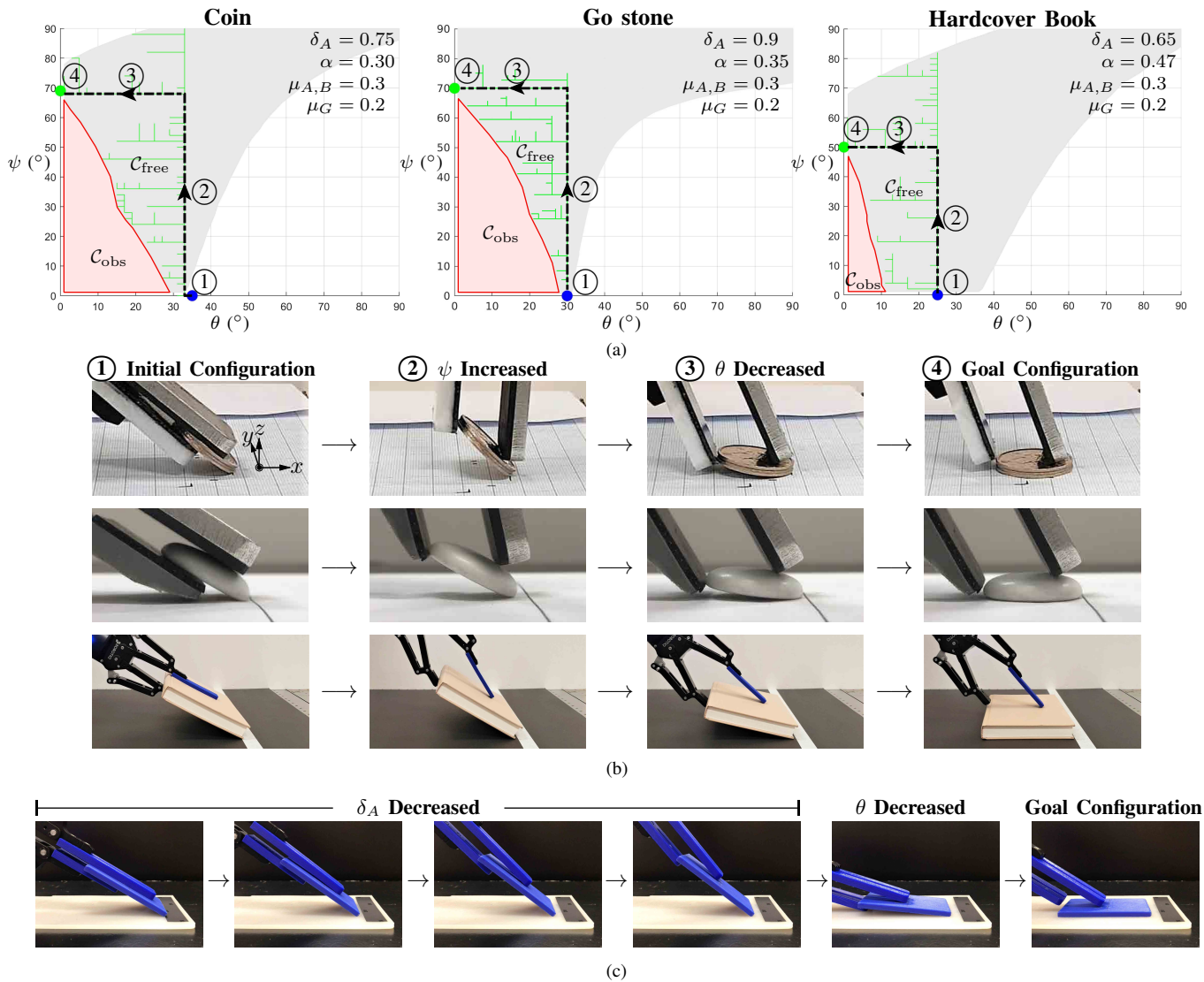
\begin{figure*}[t]
\centering 
\includegraphics[width=\textwidth]{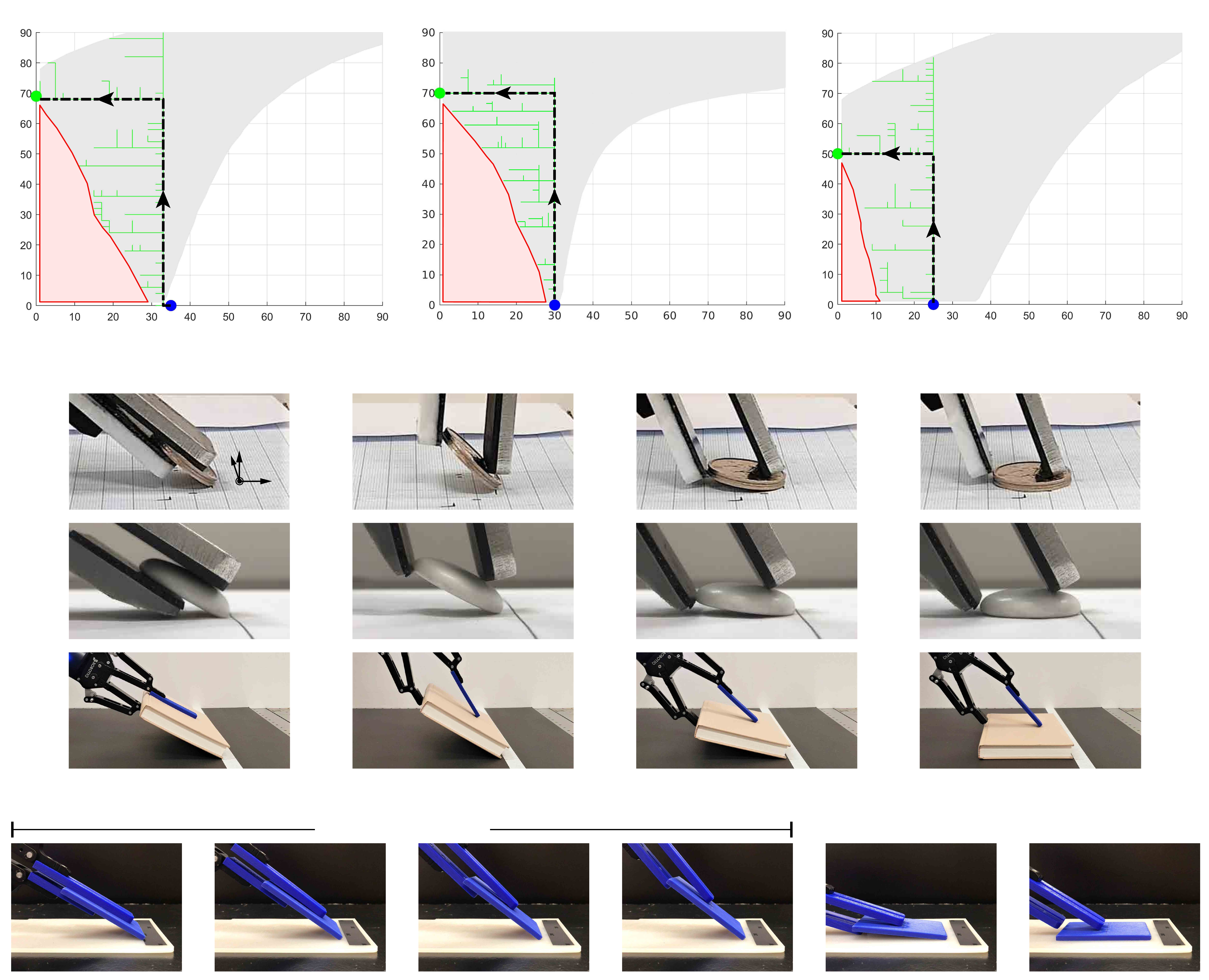}
\setlength{\unitlength}{1cm}
\begin{picture}(0,0)
\put(-6.2,14.43){\textbf{Coin}}
\put(-6.4,10.45){{\large \textcircled{\small 1}}}
\put(-7.9,10.85){{\footnotesize $\mathcal{C}_{\mathrm{obs}}$}}
\put(-7.3,12.25){{\footnotesize $\mathcal{C}_{\mathrm{free}}$}}
\put(-6.4,11.8){{\large \textcircled{\small 2}}}
\put(-7.5,13.55){{\large \textcircled{\small 3}}}
\put(-8.35,13.55){{\large \textcircled{\small 4}}}
\put(-4.6,14.12){\footnotesize $\delta_{A}=0.75$}
\put(-4.45,13.79){\footnotesize $\alpha=0.30$}
\put(-4.93,13.46){\footnotesize $\mu_{A,B}=0.3$}
\put(-4.65,13.14){\footnotesize $\mu_{G}=0.2$}
\put(-6.1,9.85){\footnotesize $\theta$ ($^\circ$)}
\put(-9.15,12.27){\footnotesize $\psi$ ($^\circ$)}

\put(-0.5,14.43){\textbf{Go stone}}
\put(-0.67,10.45){{\large \textcircled{\small 1}}}
\put(-2.0,11.2){{\footnotesize $\mathcal{C}_{\mathrm{obs}}$}}
\put(-1.5,12.43){{\footnotesize $\mathcal{C}_{\mathrm{free}}$}}
\put(-0.67,11.85){{\large \textcircled{\small 2}}}
\put(-1.7,13.63){{\large \textcircled{\small 3}}}
\put(-2.45,13.63){{\large \textcircled{\small 4}}}
\put(1.32,14.12){\footnotesize $\delta_{A}=0.9$}
\put(1.48,13.79){\footnotesize $\alpha=0.35$}
\put(1.0,13.46){\footnotesize $\mu_{A,B}=0.3$}
\put(1.28,13.14){\footnotesize $\mu_{G}=0.2$}
\put(-0.21,9.85){\footnotesize $\theta$ ($^\circ$)}
\put(-3.25,12.27){\footnotesize $\psi$ ($^\circ$)}

\put(4.7,14.4){\textbf{Hardcover Book}}
\put(4.9,10.45){{\large \textcircled{\small 1}}}
\put(3.45,10.5){{\footnotesize $\mathcal{C}_{\mathrm{obs}}$}}
\put(4.05,11.9){{\footnotesize $\mathcal{C}_{\mathrm{free}}$}}
\put(4.9,11.4){{\large \textcircled{\small 2}}}
\put(4.05,12.75){{\large \textcircled{\small 3}}}

\put(3.4,12.75){{\large \textcircled{\small 4}}}
\put(7.15,14.12){\footnotesize $\delta_{A}=0.65$}
\put(7.30,13.79){\footnotesize $\alpha=0.47$}
\put(6.82,13.46){\footnotesize $\mu_{A,B}=0.3$}
\put(7.1,13.14){\footnotesize $\mu_{G}=0.2$}
\put(5.67,9.85){\footnotesize $\theta$ ($^\circ$)}
\put(2.63,12.27){\footnotesize $\psi$ ($^\circ$)}

\put(-0.15,9.45){\footnotesize (a)}

\put(-7.98,9.15){\textbf{{\large \textcircled{\small 1}}  \small Initial Configuration}}
\put(-3.3,9.15){\textbf{{\large \textcircled{\small 2}}  \small $\psi$ Increased}}
\put(0.87,9.15){\textbf{{\large \textcircled{\small 3}}  \small $\theta$ Decreased}}
\put(4.58,9.15){\textbf{{\large \textcircled{\small 4}}  \small Goal Configuration}}
\put(-4.9,7.68){$x$}
\put(-5.45,8.25){$z$}
\put(-5.63,8.15){$y$}
\put(-4.45,8.01){$\longrightarrow$}
\put(-0.3,8.01){$\longrightarrow$}
\put(3.9,8.01){$\longrightarrow$}
\put(-4.45,6.17){$\longrightarrow$}
\put(-0.3,6.17){$\longrightarrow$}
\put(3.9,6.17){$\longrightarrow$}
\put(-4.45,4.3){$\longrightarrow$}
\put(-0.3,4.3){$\longrightarrow$}
\put(3.9,4.3){$\longrightarrow$}
\put(-0.15,3.15){\footnotesize (b)}

\put(-0.15,0.1){\footnotesize (c)}
\put(-3.9,2.55){\textbf{\small $\delta_A$ Decreased}}
\put(3.6,2.55){\textbf{\small $\theta$ Decreased}}
\put(6.15,2.55){\textbf{\small Goal Configuration}}
\put(-6.15,1.45){$\rightarrow$}
\put(-3.15,1.45){$\rightarrow$}
\put(-0.17,1.45){$\rightarrow$}
\put(2.85,1.45){$\rightarrow$}
\put(5.83,1.45){$\rightarrow$}

\end{picture}
\caption{(a) Planned ungrasping paths shown as the dashed line with the waypoints $1$-$2$-$3$-$4$ inside $\mathcal{C}_{\mathrm{free}}$ shaded gray. The path for the Go stone is the same as Fig.~\ref{fig:successful_examples}(c); here, it is projected on the plane $\delta_A=0.9$.
(b) Progress of precision placement with the objects, following the waypoints $1$-$2$-$3$-$4$. The reference frame shown in the top left panel is aligned with the one in Fig.~\ref{fig:local_geometry}(a). (c) Placement through the motion primitive $\mathsf{R}_G\mathsf{S}_A\mathsf{S}_B$. See also the video attachment.}
\label{fig:placing_new} 
\end{figure*}

\newcolumntype{A}{>{\centering\arraybackslash\hsize=.35\hsize}X}
\newcolumntype{B}{>{\centering\arraybackslash\hsize=.4\hsize}X}
\newcolumntype{D}{>{\centering\arraybackslash\hsize=.4\hsize}X}
\newcolumntype{E}{>{\centering\arraybackslash\hsize=.35\hsize}X} 
\newcolumntype{F}{>{\centering\arraybackslash\hsize=.15\hsize}X}
\newcolumntype{G}{>{\centering\arraybackslash\hsize=.2\hsize}X}
\newcolumntype{H}{>{\centering\arraybackslash\hsize=.2\hsize}X}
\newcolumntype{I}{>{\centering\arraybackslash\hsize=.2\hsize}X}
\newcolumntype{J}{>{\centering\arraybackslash\hsize=.2\hsize}X}

\begin{table*}
\caption{Experiment results of precision placement.}
\centering 
\begin{tabularx}{\textwidth}{@{}ABDEFGHIJ@{}} 
\hline\hline
\cr
\multirow{2}{*}{\shortstack[c]{Section\\Reference}} & \multirow{2}{*}{\shortstack[c]{Task\\Object}} & {\shortstack[c]{Object\:Dimension\\$\ell_{\mathrm{obj}} \times$ thickness}} & {\shortstack[c]{Digit\\Asymmetry}} & \multicolumn{2}{p{2.3cm}}{\centering {\shortstack[c]{Initial\\Configuration}}} & \multicolumn{2}{p{2.3cm}}{\centering{\shortstack[c]{Placement\\Error}}} & \multirow{2}{*}{\shortstack[c]{Success\\Rate}} \\
\cmidrule(lr){5-6}\cmidrule(lr){7-8}
 &  & (mm) & $\alpha$ & $\delta_{A}$ & $\theta$ ($^\circ$) & Mean\:(mm) & Range\:(mm) &  \\[0.5ex]

\hline
\cr
\multirow{4}{*}{{\shortstack[c]{\vspace{4pt}\\\ref{subsec:basic_placement}\\Digit\\Asymmetry}}}  
& Coin &  $27\times3$ & $[0.20,0.30]$ & 0.75 & $[30,35]$ & $-1.1$ & $[-3, 1]$ & 53/60\\[0.5ex]
& Go stone &  $23\times5.7$ & $[0.22,0.35]$ & 0.9 & $[30,35]$ & $1.7$ & $[-1, 4]$ & 55/60\\[0.5ex]
& Hardcover book &  $160\times25$ & $[0.42,0.47]$ & 0.65 & $[20,35]$ &$0.9$ & $[-1, 3]$ & 51/60\\[0.5ex]
& 3D-printed plate$^{1}$ & $85\times5$ & 0.41 & 0.79 & 30 & $-3.1$ & $[-5,-1]$ & 37/40\\[0.5ex]
\hline 
\rule{0 pt}{2.5ex}
\multirow{3}{*}{\shortstack[c]{\vspace{1pt}\\\ref{subsec:SF}\\Same-Length\\Fingers}} 
& Coin & $27\times3$ & 0.0 & 0.75 & 35 & $10.9$ & $[6,17]$ & 2/20\\[0.5ex]
& Go stone & $23\times5.7$ & 0.0 & 0.9 & 35 & $8.3$ & $[5,11]$ & 3/20\\[0.5ex]
& Hardcover book & $160\times25$ & 0.0 & 0.65 & 25 & $62.7$ & $[51,73]$ & 0/20\\[0.5ex]
\hline\hline
\end{tabularx}
\rule{0pt}{2ex}
{\raggedright 
1 Placed onto a rubber mat with high friction.\\
}
\label{table:results} 
\end{table*}

\subsection{Placement with Digit Asymmetry}
\label{subsec:basic_placement}
\begin{figure}[H]
\centering 
\includegraphics[width=\columnwidth]{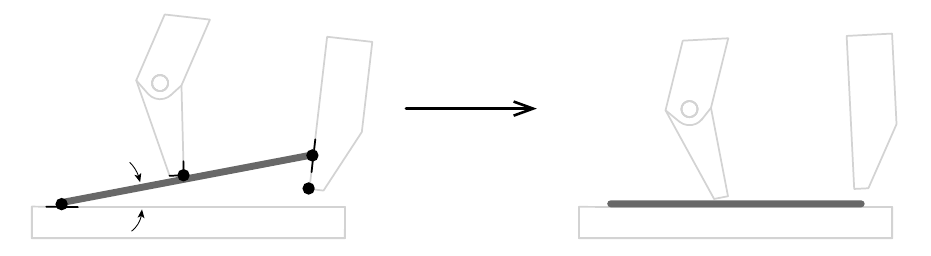}
\setlength{\unitlength}{1cm}
\begin{picture}(0,0)
\put(-2.87,1.23){\footnotesize $A$}
\put(-1.73,0.95){\footnotesize $T$}
\put(-3.25,1.3){\footnotesize $\theta$}
\put(-3.2,2.3){\footnotesize \rotatebox{30}{Finger}}
\put(-1.4,2.2){\footnotesize \rotatebox{30}{Thumb}}
\end{picture}
\caption{Push-down maneuver to complete ungrasping while avoiding collision at the thumb. The fingertip at $A$ pushes the object down as the gripper is controlled to rotate about the thumb's tip $T$.}
\label{fig:engage} 
\end{figure}
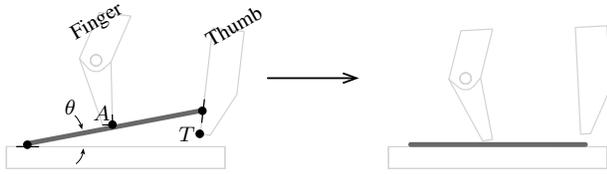

Given object geometry and friction properties, the ungrasping paths for the coin, the Go stone, and the book using the two motion primitives $\mathsf{R}_G\mathsf{R}_A\mathsf{S}_B$ and $\mathsf{R}_G\mathsf{R}_A\mathsf{R}_B$ were planned and executed in an open-loop manner, 60 times for each object by varying the choice of the digit asymmetry $\alpha$ and initial $\theta$ (the angle between the object and the target surface). 
Instances of the planned paths are presented in Fig.~\ref{fig:placing_new}(a), along the waypoints $1$-$2$-$3$-$4$. Initially, the object was pinch-grasped from a stand as shown in Fig.~\ref{fig:exp_setting}. 
Before terminating an ungrasp, at a small value of $\theta$, the robot is controlled to push the object down (Fig.~\ref{fig:engage}) by rotating the gripper about the thumb's tip in order to keep a safety distance between the thumb and the target surface. The gripper's internal mobility was simply locked during this push-down maneuver, considering that its duration is short and the gripper has some passive compliance. 
Fig.~\ref{fig:placing_new}(b) shows the progress for each object. Table~\ref{table:results} reports two measures: success rate and placement error. An attempt of placement is regarded as a success if no loss of grasp occurs until the final push-down maneuver (Fig.~\ref{fig:engage})
is performed. The overall success rate was then 159/180 (successes/attempts). In the failed attempts, the contact forces by the gripper were unable to counterbalance the weight of the object, and thus premature loss of grasp happened. This may account for why the heaviest object, the hardcover book, had the lowest success rate. The average placement errors, measured as the final positioning error of the distal end of the object on the surface along the $x$-axis shown in Fig.~\ref{fig:placing_new}(b), were $-1.1$mm, $1.7$mm, and $0.9$mm for the coin, the Go stone, and the hardcover book, respectively. 

Fig.~\ref{fig:placing_new}(c) shows ungrasping with the 3D-printed plate when the motion primitive $\mathsf{R}_G\mathsf{S}_A\mathsf{S}_B$ was applied. Because the gripper exerts pulling forces at both $A$ and $B$, a feasible motion path is obtained with large friction at $G$. Still, the average placement error was relatively large at $-3.1$mm.

\subsection{Placement with Same-Length Fingers}
\label{subsec:SF}
It is not possible to find feasible paths with same-length fingers ($\alpha=0$) because $\mathbf{q}_\mathrm{goal}$ where $\theta=0^\circ$ and $\psi=90^\circ$ lies outside $\mathcal{C}_{\mathrm{free}}$, as can be seen in Fig.~\ref{fig:placing_new}(a). When the placement attempts were made anyway, the overall success rate was significantly lower at 5/60.
As also reported in the table, placement errors significantly surged as the object dynamically slid out from the gripper.

\section{Conclusion}
\label{sec:discussion_conclusion}

We presented a formulation for ungrasping through dexterous manipulation. Our planning framework based on a sampling-based search algorithm makes it possible to find an optimal course of secure, feasible manipulation actions specifying the necessary contact interactions.
The reported experiments show that our current implementation of dexterous ungrasping is capable of addressing the novel challenges in placing real-world objects through dexterous manipulation. Our practical design feature, i.e. digit asymmetry, proved its effectiveness in securely holding the object through the experiments, as well as the numerical examples. The model of ungrasping manipulation established with the simple gripper model allows the use of a wide variety of grippers, apart from the parallel-finger gripper used in the experiments.

\bibliographystyle{ieeetr}
\bibliography{references}

\begin{thebibliography}{10}

\bibitem{897777}
A.~{Bicchi}, ``Hands for dexterous manipulation and robust grasping: a
  difficult road toward simplicity,'' {\em IEEE Transactions on Robotics and
  Automation}, vol.~16, no.~6, pp.~652--662, 2000.

\bibitem{mason1985robot}
M.~T. Mason and J.~K. Salisbury~Jr, ``Robot hands and the mechanics of
  manipulation,'' 1985.

\bibitem{1087063}
R.~{Fearing}, ``Simplified grasping and manipulation with dextrous robot
  hands,'' {\em IEEE Journal on Robotics and Automation}, vol.~2, no.~4,
  pp.~188--195, 1986.

\bibitem{rus1999hand}
D.~Rus, ``In-hand dexterous manipulation of piecewise-smooth 3-d objects,''
  {\em The International Journal of Robotics Research}, vol.~18, no.~4,
  pp.~355--381, 1999.

\bibitem{han1998dextrous}
L.~Han and J.~C. Trinkle, ``Dextrous manipulation by rolling and finger
  gaiting,'' in {\em Proceedings. 1998 IEEE International Conference on
  Robotics and Automation}, pp.~730--735, IEEE, 1998.

\bibitem{1087910}
P.~{Tournassoud}, T.~{Lozano-Perez}, and E.~{Mazer}, ``Regrasping,'' in {\em
  Proceedings. 1987 IEEE International Conference on Robotics and Automation},
  vol.~4, pp.~1924--1928, 1987.

\bibitem{trinkle1990planning}
J.~C. Trinkle and R.~P. Paul, ``Planning for dexterous manipulation with
  sliding contacts,'' {\em The International Journal of Robotics Research},
  vol.~9, no.~3, pp.~24--48, 1990.

\bibitem{cole1992dynamic}
A.~A. Cole, P.~Hsu, S.~S. Sastry, {\em et~al.}, ``Dynamic control of sliding by
  robot hands for regrasping,'' {\em IEEE Transactions on robotics and
  automation}, vol.~8, no.~1, pp.~42--52, 1992.

\bibitem{montana1988kinematics}
D.~J. Montana, ``The kinematics of contact and grasp,'' {\em The International
  Journal of Robotics Research}, vol.~7, no.~3, pp.~17--32, 1988.

\bibitem{614264}
L.~{Han}, Y.~S. {Guan}, Z.~X. {Li}, Q.~{Shi}, and J.~C. {Trinkle}, ``Dextrous
  manipulation with rolling contacts,'' in {\em Proceedings of International
  Conference on Robotics and Automation}, April 1997.

\bibitem{yuanicra2020}
S.~Yuan, A.~D. Epps, J.~B. Nowak, and J.~K. Salisbury, ``Design of a
  roller-based dexterous hand for object grasping and within-hand
  manipulation,'' in {\em 2020 International Conference on Robotics and
  Automation (ICRA)}, IEEE, 2020.

\bibitem{6907062}
N.~C. {Dafle}, A.~{Rodriguez}, R.~{Paolini}, B.~{Tang}, S.~S. {Srinivasa},
  M.~{Erdmann}, M.~T. {Mason}, I.~{Lundberg}, H.~{Staab}, and T.~{Fuhlbrigge},
  ``Extrinsic dexterity: In-hand manipulation with external forces,'' in {\em
  2014 IEEE International Conference on Robotics and Automation (ICRA)},
  pp.~1578--1585, May 2014.

\bibitem{7913727}
J.~{Shi}, J.~Z. {Woodruff}, P.~B. {Umbanhowar}, and K.~M. {Lynch}, ``Dynamic
  in-hand sliding manipulation,'' {\em IEEE Transactions on Robotics}, vol.~33,
  no.~4, pp.~778--795, 2017.

\bibitem{deimel2016novel}
R.~Deimel and O.~Brock, ``A novel type of compliant and underactuated robotic
  hand for dexterous grasping,'' {\em The International Journal of Robotics
  Research}, vol.~35, no.~1-3, pp.~161--185, 2016.

\bibitem{doi:10.1177/0278364918802346}
V.~Babin and C.~Gosselin, ``Picking, grasping, or scooping small objects lying
  on flat surfaces: A design approach,'' {\em The International Journal of
  Robotics Research}, vol.~37, no.~12, pp.~1484--1499, 2018.

\bibitem{mason2001mechanics}
M.~T. Mason, {\em Mechanics of robotic manipulation}.
\newblock MIT press, 2001.

\bibitem{simunovic1975force}
S.~Simunovic, ``Force information in assembly processes,'' in {\em 5th
  International Symposium on Industrial Robots}, pp.~415--431, 1975.

\bibitem{watson1978remote}
P.~C. Watson, ``Remote center compliance system,'' July~4 1978.
\newblock US Patent 4,098,001.

\bibitem{shome2019towards}
R.~Shome, W.~N. Tang, C.~Song, C.~Mitash, H.~Kourtev, J.~Yu, A.~Boularias, and
  K.~E. Bekris, ``Towards robust product packing with a minimalistic
  end-effector,'' in {\em 2019 International Conference on Robotics and
  Automation (ICRA)}, pp.~9007--9013, IEEE, 2019.

\bibitem{9384169}
R.~Newbury, K.~He, A.~Cosgun, and T.~Drummond, ``Learning to place objects onto
  flat surfaces in upright orientations,'' {\em IEEE Robotics and Automation
  Letters}, vol.~6, no.~3, pp.~4377--4384, 2021.

\bibitem{8598749}
C.~H. {Kim} and J.~{Seo}, ``Shallow-depth insertion: Peg in shallow hole
  through robotic in-hand manipulation,'' {\em IEEE Robotics and Automation
  Letters}, vol.~4, pp.~383--390, April 2019.

\bibitem{Lynch:2017:MRM:3165183}
K.~M. Lynch and F.~C. Park, {\em Modern Robotics: Mechanics, Planning, and
  Control}.
\newblock New York, NY, USA: Cambridge University Press, 2017.

\bibitem{doi:10.1177/0278364919880257}
N.~Chavan-Dafle, R.~Holladay, and A.~Rodriguez, ``Planar in-hand manipulation
  via motion cones,'' {\em The International Journal of Robotics Research},
  vol.~39, no.~2-3, pp.~163--182, 2020.

\bibitem{doi:10.1177/0278364911406761}
S.~Karaman and E.~Frazzoli, ``Sampling-based algorithms for optimal motion
  planning,'' {\em The International Journal of Robotics Research}, vol.~30,
  no.~7, pp.~846--894, 2011.

\end{thebibliography}

\end{document}